\documentclass[letterpaper, 10 pt, conference]{ieeeconf}  %

\IEEEoverridecommandlockouts                              %

\usepackage{times}
\usepackage{graphics} %
\usepackage{graphicx}
\usepackage{subfigure}
\usepackage{amsmath,amssymb,amsopn,amstext,amsfonts}
\usepackage[ruled,vlined]{algorithm2e}
\SetKwComment{Comment}{$\triangleright$\ }{}
\usepackage{cancel}
\usepackage[space]{cite}
\usepackage{pdfsync}
\usepackage{balance}
\usepackage{color}
\usepackage{mathtools}
\usepackage{bm}

\usepackage{diagbox}
\usepackage{float}
\usepackage{epstopdf}
\usepackage{pifont}
\usepackage{fixltx2e}
\usepackage{amsmath}
\usepackage{multirow}
\usepackage{url}
\usepackage{verbatim}
\usepackage{caption}
\usepackage{adjustbox}
\usepackage{booktabs}
\usepackage{threeparttable}
\usepackage{makecell}
\usepackage{setspace}
\usepackage{array}
\usepackage{mdframed}
\usepackage{frame}
\usepackage{framed}
\usepackage{tabularx}
\makeatletter
\let\NAT@parse\undefined
\makeatother
\usepackage[linkcolor=red,citecolor=green,urlcolor=green,colorlinks=true]{hyperref}
\title{\LARGE \bf
Event-based Motion Segmentation by Cascaded \\ Two-Level Multi-Model Fitting}

\author{Xiuyuan Lu, Yi Zhou*, and Shaojie Shen%
\thanks{All authors are with the Dept. of ECE, Hong Kong University of Science and Technology, Hong Kong, China. Email: $\{$xluaj, eeyzhou, eeshaojie$\}$@ust.hk.
This work was supported by the HKPFS and HKUST PDF matching fund 2020.
*Corresponding author.}
}

\global\long\def\bx{\mathbf{x}}

\global\long\def\Rot{\mathtt{R}}

\global\long\def\cT{\mathcal{T}}

\global\long\def\cS{\mathcal{S}} %
\global\long\def\cF{\mathcal{F}} %
\global\long\def\cS{\mathcal{S}} %

\global\long\def\cM{\mathcal{M}}
\global\long\def\cL{\mathcal{L}}

\global\long\def\bm{\mathbf{m}} %
\global\long\def\cE{\mathcal{E}} %
\global\long\def\cS{\mathcal{S}} %

\begin{document}

\maketitle
\thispagestyle{empty}
\pagestyle{empty}

\begin{abstract}

Among prerequisites for a synthetic agent to interact with dynamic scenes, the ability to identify independently moving objects is specifically important.
From an application perspective, nevertheless, standard cameras may deteriorate remarkably under aggressive motion and challenging illumination conditions.
In contrast, event-based cameras, as a category of novel biologically inspired sensors, deliver advantages to deal with these challenges.
Its rapid response and asynchronous nature enables it to capture visual stimuli at exactly the same rate of the scene dynamics.
In this paper, we present a cascaded two-level multi-model fitting method for identifying independently moving objects (i.e., the motion segmentation problem) with a monocular event camera.
The first level leverages tracking of event features and solves the feature clustering problem under a progressive multi-model fitting scheme.
Initialized with the resulting motion model instances, the second level further addresses the event clustering problem using a spatio-temporal graph-cut method.
This combination leads to efficient and accurate event-wise motion segmentation that cannot be achieved by any of them alone.
Experiments demonstrate the effectiveness and versatility of our method in real-world scenes with different motion patterns and an unknown number of independently moving objects. 
\end{abstract}

\section{Introduction}
\label{sec: introduction}

\begin{figure}[t]
\vspace{0.9cm}
  \centering
  \captionsetup{skip=1ex}
  \subfigure[\small{Raw intensity image from the standard camera.}]{
  \frame{\includegraphics[width=0.45\columnwidth]{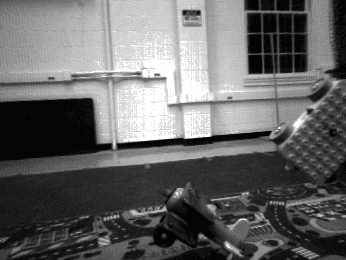}}}
  \subfigure[\small{Event feature extraction and tracking.}]{
  \frame{\includegraphics[width=0.45\columnwidth]{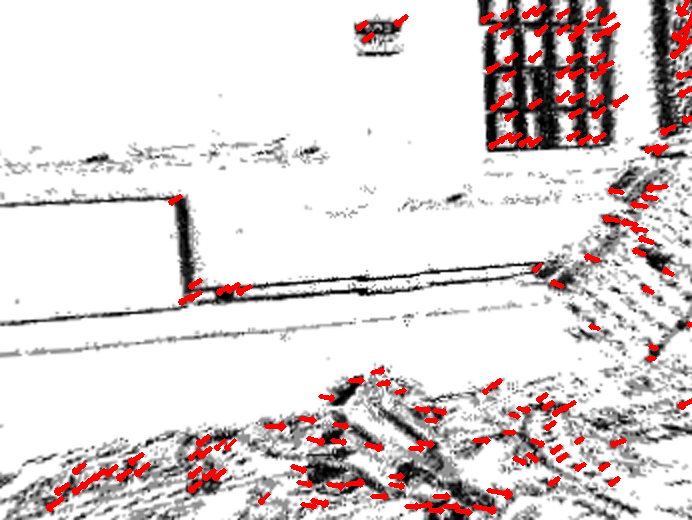}}}
  \subfigure[\small{Event feature clustering.}]{
  \frame{\includegraphics[width=0.45\columnwidth]{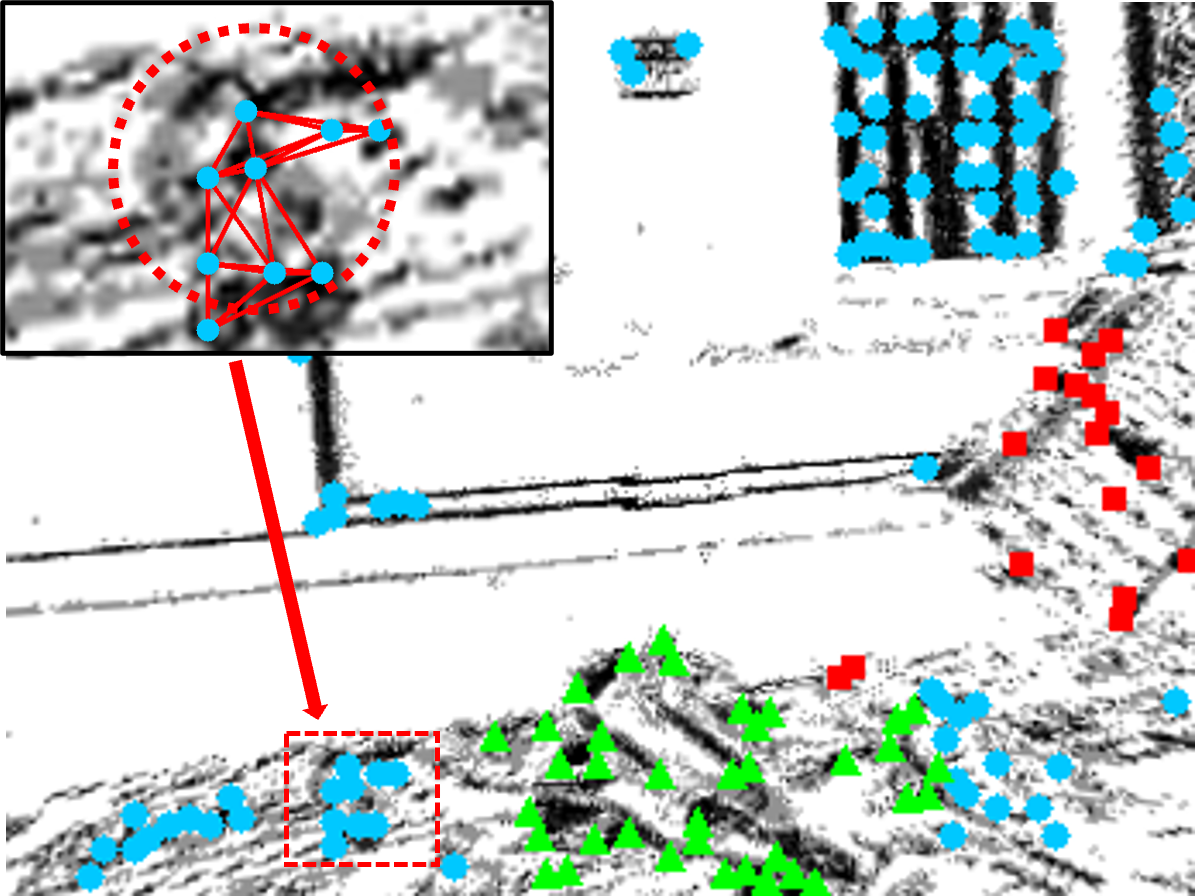}}}
  \subfigure[\small{Event clustering and alignment.}]{
  \frame{\includegraphics[width=0.45\columnwidth]{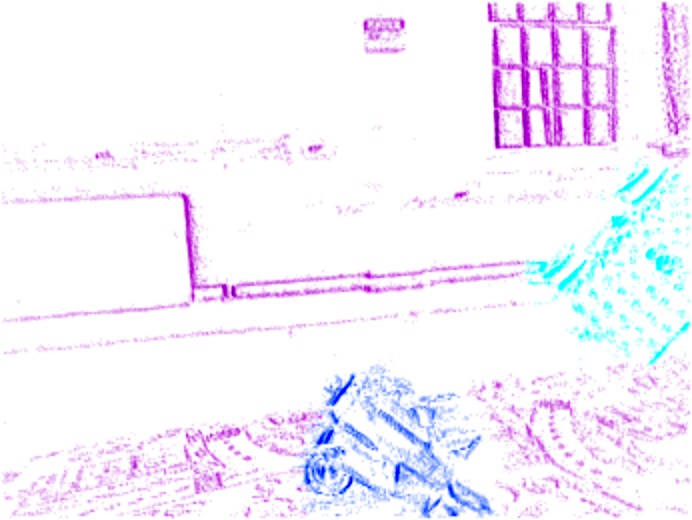}}}
 \caption{The proposed cascaded two-level multi-model fitting scheme for event clustering. 
 The raw intensity image (a) is for visualization only.
 Event features are extracted on the uncompensated image of warped events and continuously tracked on the image plane (b).
 At level one (c), event features are clustered into groups that undergo different rigid motions.
 At level two (d), events are clustered by associating to the refined motion models and are aligned along corresponding point trajectories.}
 \label{fig:eye-catcher-four-images}
 \vspace{-0.5cm}
 \end{figure}%
\begin{figure*}[t]
  \centering
  \includegraphics[width=0.95\linewidth]{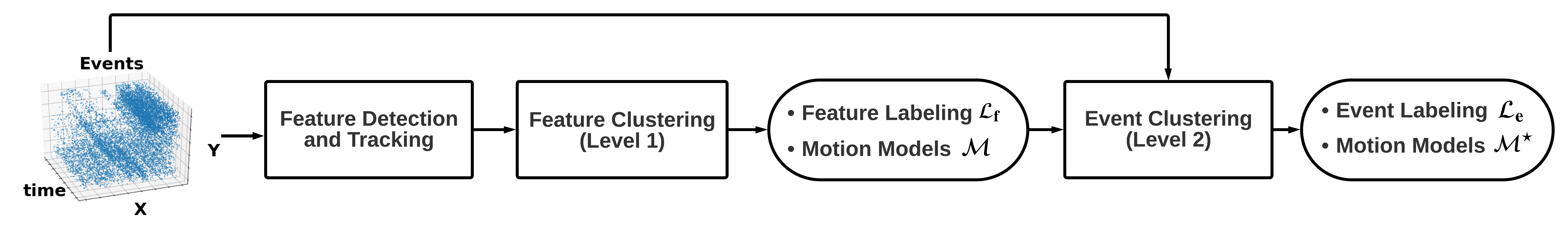}
   \captionof{figure}{Overview of the proposed system. 
The input to the system is a stream of events occurred within a short time interval.
The proposed system consists of two multi-model fitting modules cascaded sequentially.
Using as input the event features' tracking, level one clusters features (labeled by $\cL_\text{f}$) into groups that comply with different rigid motions $\cM$.
Initialized by level one, events are clustered (labeled by $\cL_\text{e}$) at level two by associating with refined motion models $\cM^{\star}$.}
  \label{fig:flowchart}
\end{figure*}

Modern robotic applications (e.g., autonomous driving, AR/VR, UAVs, etc) require synthetic agents to perceive dynamic scenes in order to interact with the environment effectively.
As one of the challenges cast by dynamic scenes, the multi-motion segmentation problem aims at detecting and further modeling the motions of independently moving objects (IMOs) while estimating the camera's ego-motion.
The problem's complexity originates from its exploratory nature: multiple hypotheses (models) are to be considered and validated against the data to select the optimal fits.
Moreover, from a sensing perspective, challenging visual effects (e.g., motion blur and over/under-exposure) make the problem even harder for solutions using standard cameras.

Event-based cameras, for example, the DVS~\cite{Lichtsteiner08ssc}, are biologically-inspired visual sensors that mimic the transient pathway of the human visual system.
They acquire visual information in a completely different way from standard cameras, and consequently, produce a stream of asynchronous per-pixel intensity changes (the so-called "events'') instead of intensity images.
This novel principle of operation brings advantages over standard frame-based cameras to address the %
challenging scenarios in robotic vision, such as high-speed and/or high dynamic range (HDR) stereo depth estimation \cite{Rogister12tnnls, Zhou18eccv}, camera tracking \cite{Gallego17pami, Mueggler14iros} and Simultaneous Localization and Mapping \cite{Kim16eccv, Rebecq17ral, zhou2021event}.

In this paper, we look into the problem of event-based motion segmentation, which aims at clustering events occurred within a short time interval into groups, such that each of them complies with a unique rigid motion (background motion or an IMO's motion).
A novel method is developed that jointly clusters the events (data-model association) and estimates the parameters of their coherent motion (model fitting).
The contribution of the work is summarized as:
\begin{itemize}
    \item A cascaded two-level multi-model fitting scheme that solves the event-based motion segmentation problem by two steps: event feature clustering (level one) and event clustering (level two).
    \item An efficient hypothesis proposing unit (level one) developed by integrating a robust model fitting method into a progressive multi-model fitting pipeline.
    The result of level one -- a compact set of motion instances -- is used to initialize the process in level two, which leads to efficient event-wise labeling.
    \item An extensive evaluation on available datasets showing state-of-the-art performance.
\end{itemize}

In the rest of the paper, we first review related work (Section~\ref{sec: related work}), and then explain our method (Section~\ref{sec:problem} and \ref{sec:method}).
Finally, the proposed approach is evaluated in Section~\ref{sec:evaluation}.

\section{Related Work}
\label{sec: related work}
The literature review focuses on works that do not require prior knowledge, including the shape of IMOs, and the correlation between the tracked geometric primitives and the event camera's motion.
Early prior-free pipelines (e.g., \cite{Mitrokhin18iros, Stoffregen17acra}) follow a greedy strategy, which first recover the dominant background motion and deal with the remaining events induced by the IMOs subsequently.

Using the result of \cite{Stoffregen17acra} as initialization, \cite{Stoffregen19iccv} presented the first method that solved the model estimation (fitting) and event-motion association (labeling) sub-problems jointly.
The segmentation problem was formulated in an expectation-maximization (EM) scheme, which iteratively switched between updating the event-motion association and refining the motion models.
Both the E-step and M-step leveraged the idea of motion compensation~\cite{Gallego18cvpr}.
The method, however, required to know the number of IMOs as a prior knowledge.

More recently, a hierarchical clustering method was presented in~\cite{parameshwara2020moms} and further improved in~\cite{parameshwara20200mms}.
The method first clustered event features by applying the K-means method.
Then the resulting redundant clusters were passed to an iterative method, which consisted of clusters merging and motion parameters refinement.
The method does not require to know the number of IMOs in advance (by setting $K$ to a big number).
However, the clusters merging criteria was somehow designed under a hybrid metric of image contrast and centroid distance.
Thus, clusters undergoing identical motions could not be merged if their centroids were far away from each other.
Finally, the event-motion association (event-wise labeling) was made according to if an event was within the convex hull of the features in a cluster.
Inaccurate event labeling was witnessed near the overlapping areas of background and IMOs.

To address the flaws of \cite{Stoffregen19iccv}\cite{parameshwara2020moms}, the method in~\cite{zhou2020eventbased} adapted motion compensation \cite{Gallego18cvpr} to a graph-based energy formulation.
The idea of negative representation \cite{zhou2021event} is borrowed so that the original maximization problem is converted to a minimization one.
The number of clusters could converge to the real number of IMOs because of the applied constraint on the length of description.
However, this method failed to detect IMOs with a small size because of the bad inlier-outlier ratio in the initialization.
In addition to the above methods, we have to mention a supervised end-to-end learning-based pipeline in~\cite{Mitrokhin19iros}, which jointly estimated optical flow, 3D motion and object segmentation. 
It provided a state-of-the-art dataset for evaluation of IMOs segmentation, though it was not close to any other approaches in the literature review.

Our method draws on the advantages of existing methods and overcomes their shortcomings with a new design.
Inspired by \cite{zhou2020eventbased}, we also formulate the event-based segmentation as a geometric multi-model fitting problem.
Unlike \cite{zhou2020eventbased}, we propose a cascaded two-level multi-model fitting scheme.
In the first level, we bring in event feature tracking \cite{rodriguez2020asynchronous, parameshwara2020moms, parameshwara20200mms} for efficient hypothesis proposal.
Instead of using K-means \cite{parameshwara2020moms, parameshwara20200mms}, we cluster the features with a progressive multi-model fitting method.
The model fitting and feature-motion association sub-problems are solved jointly, thus not requiring incremental cluster merging locally~\cite{parameshwara2020moms, parameshwara20200mms}.
In the second level, we leverage the spatio-temporal graph cut method \cite{zhou2020eventbased} to solve the event clustering problem.

\section{Problem Statement and Preliminaries}
\label{sec:problem}

The goal is to cluster events occurred within a short time interval into several groups such that each complies with a unique rigid motion.
Each motion model parametrizes the 2D point trajectories of events on the image plane induced by an IMO.
Once the events are clustered, they can be warped according to the estimated motions, thus producing an image of warped events with the highest contrast (Fig.~\ref{fig:eye-catcher-four-images} (d)).

In this section, we first review the principles of event cameras and clarify the input of our algorithm.
Second, we discuss the motion model fitting methods using as input the feature correspondences and the raw events, respectively.
Finally, the event-based motion segmentation problem is identified as a geometric multi-model fitting one, followed by an overview of the proposed cascaded two-level multi-model fitting scheme.

\subsection{Event Stream and Features}
\label{subsec:event stream and features}

Unlike standard cameras which capture full frames at a fixed frame rate, event cameras have independent pixels that sense brightness changes asynchronously.
An event is triggered if the variation of the sensed intensity (in log scale) $I(\bx,t)$, at a pixel location $\bx = (u,v)^{\text{T}}$ and time instance $t$, exceeds a nominal threshold $C_{\text{th}}$:
\begin{equation}
\label{eq:DVS principle}
    \vert \log{I(\bx,t)} - \log{I(\bx, t-\Delta t)} \vert > C_{\text{th}},
\end{equation}
where $\Delta t$ is the time since the last event was triggered at this pixel.
Each event $e$, denoted by a tuple $<u,v,t,p>$, records spatio-temporal information that includes $u, v$ the pixel coordinate of the event, $t$ the timestamp at which the event occurs, and $p \in \{-1, +1\}$ the polarity indicating the sign of the brightness change.

We denote the input of our algorithm, a stream of events occurred within a short time interval $[t-\delta t, t]$, as $\cE_{[t-\delta t,t]} = \{e_k\}_{k=1}^{N_e}$.
Besides, we also use the event feature correspondences as input.
We extract event features %
at time instance $t-\delta t$ and $t$, respectively, and establish feature correspondences by tracking the 2D motion of these features (detailed in Section~\ref{subsec: level one}).
The set of feature correspondences is denoted by $\cF = \{\mathbf{f}_i\}_{i=1}^{N_\mathbf{f}}$,
where $\mathbf{f}_i = \{ \bx_i^{t-\delta t} \leftrightarrow \bx_i^{t} \}$ refers to a pair of feature correspondence.

\subsection{Motion Model Fitting}
\label{subsec:motion model fitting}
Here we clarify the ways to fit a motion model with feature correspondences and raw events, respectively.
At level one of our method, motion models can be estimated in a closed form given feature correspondences.
Using the four-parameter model $\mathbf{m}:=\{m_u, m_v, m_s, m_{\theta}\}^{\text{T}}$~\cite{Mitrokhin18iros} as an example, the point trajectory on the image plane is parametrized as:
\begin{equation}
\label{eq:four_param_motion_model}
\bx^{\prime}_1 = \bx_1 +   
    \left[ 
        \begin{matrix}    
            \left(\begin{matrix}    
                m_u  \\    m_v \\   
            \end{matrix}  \right)  + (m_s + 1) \Rot_{m_{\theta}} \bx_1 - \bx_1 
    \end{matrix}  \right]\delta t, 
\end{equation}
where $\bx_1 \leftrightarrow \bx^{\prime}_1$ denotes a pair of corresponding features, $\Rot_{m_{\theta}} := \left( \begin{matrix}    \cos{m_{\theta}} &-\sin{m_{\theta}} \\ \sin{m_{\theta}} & \cos{m_{\theta}} \\ \end{matrix} \right)$ the in-plane rotation, and $\delta t$ the time interval during which the features are tracked.
The minimal case of Eq.~\ref{eq:four_param_motion_model} requires two pairs of feature correspondences, which can be solved by the direct linear transform (DLT) method.
Other models, such as 2D-flow model~\cite{Zhu17icra}, and 3D-Rotation model~\cite{Gallego17ral}, can be worked out similarly.
For a single-instance case in presence of outliers, a robust solution typically uses a minimal-set solver inside a RANSAC framework~\cite{Fischler81cacm} and subsequently refines the estimate via a nonlinear optimization with all inliers.

In the second level, the resulting motion instances by the first level will be refined with all associated events.
We apply the \textit{motion compensation}~\cite{Gallego18cvpr} method for model fitting with raw events.
The idea of \textit{motion compensation} refers to the process of warping all involved events with an associated motion model to a reference time instance.
During the process, events are aligned incrementally along the point trajectories on the image plane, and the best fit is found when the contour strength (contrast) of the image of warped events (IWE) is maximized.
The contour strength can be assessed by a variety of dispersion metrics. 
Here, we utilize the variance loss because of its advantages in terms of accuracy and computation complexity over other alternatives~\cite{Gallego19cvpr}.
The motion compensation method can be summarized as follows.
First, all involved events $\cE$ are geometrically transformed according to a warping function $\mathbf{W}$,
\begin{equation}
\label{eq:warped-events}
e_k \doteq (\bx_k, t_k) \;\mapsto\; e_k^{\prime} \doteq (\bx_k^{\prime}, t_{\text{ref}}),
\end{equation}
resulting in a number of warped events $\cE' = \{e_k^{\prime}\}_{k=1}^{N_e}$ at a reference time $t_{\text{ref}}$.
The warping function $\mathbf{W}$ is parametrized by the motion model $\textbf{m}$. 
It determines the image-plane trajectories along which the events are aligned.
Second, the warped events $\cE'$ are aggregated into an IWE,
\begin{equation}
\label{eq:IWE}
    I(\bx; \bm) \doteq \sum_{k=1}^{N_e}\delta (\bx - \bx_{k}^{\prime}(\bm)),
\end{equation}
where each pixel $\bx$ counts the number of warped events that fall within it.
To establish a differentiable objective in the contrast maximization, the Dirac function $\delta(\cdot)$ is typically replaced by a Gaussian $\mathcal{N}(\bx; \mathbf{0},\epsilon^2\text{Id})$ of $\epsilon=1$ pixel width.
Finally, the variance of the IWE quantifies the goodness of fitting: $\bm^\ast = \arg\max_{\bm} \sigma^2(I(\bx; \bm))$.
Our formulation supports any type of parametric motion model.

\subsection{Segmentation via Geometric Multi-Model Fitting}
\label{subsec:cascaded multi model fitting}

Segmentation refers to the problem of clustering data according to some proximity in data, such as identical color and texture pattern (photometric proximity) or being consistent with a common geometric model (geometric proximity).
Event-based motion segmentation lies in the latter case because the absolute intensity is not available in raw data.
The problem can be posed as an energy-based multi-model fitting problem, which jointly solves data-model association and model fitting sub-problems.
The data-model association sub-problem, also known as labeling, is a discrete optimization problem.
To solve this sub-problem, a graph structure is typically established on the data, which introduces additional consideration on spatial regularity, and more importantly, enables to apply modern graph-cut solvers.
The overall energy has a unified form $E(L, \cM) = E_\text{data}(L, \cM) + E_\text{reg}(L, \cM)$, which is a function of two sets of variables: $L$ the labeling function that associates a model to a data, and $\cM$ the models that represent the commonality shared within each cluster.

We follow this formulation in the sense of energy-based multi-model fitting with a graph structure.
Furthermore, we propose a cascaded two-level multi-model fitting scheme.
In each level, the motion segmentation problem is formulated as a joint optimization problem over the data-model (features-motion in the first level, and events-motion in the second level) association and the motion models' parameters.
The data-model association is represented by the labeling function $L(d\footnote{$d$ refers to a data which could be either an pair of corresponding features $\mathbf{f} \in \Omega \times \Omega$ or an event $e \in \Omega \times \cT$.}) : \mathbb{D} \rightarrow \cL = \{1,...,N\}$, which associates a data $d$ to a label $l \in \cL$ indicating which IMO the data belongs to.
The motion patterns of IMOs are represented by a set of motion models $\cM = \{\mathbf{m}_1,...,\mathbf{m}_N\}$.
Each motion model encodes the coherent motion that a cluster of data complies with.
The overall system is illustrated in Fig.~\ref{fig:flowchart}.
We detail the multi-model fitting method at each level in Section.~\ref{sec:method}.
\section{Methodology}
\label{sec:method}
In this section, we detail the proposed multi-model fitting method at each level.
First, we discuss the approach for feature clustering at level one, which incorporates robust motion estimation into a progressive multi-model fitting scheme.
Using the resulting motion instances as an initial model pool, we show in the second level that the event clustering problem can be efficiently addressed with a spatio-temporal graph cut method.

\subsection{Feature Clustering by Progressive Multi-Model Fitting}
\label{subsec: level one}

The multi-model fitting method, at level one, jointly solves the motion parameters estimation and feature-motion association sub-problems, using the feature correspondences as input.
We extract Shi-Tomasi corners~\cite{Shi94cvpr} on raw IWEs (without motion compensation\footnote{We observed no apparent difference in performance whether or not a global motion compensation is performed as pre-processing~\cite{parameshwara20200mms,parameshwara2020moms}.}) at time $t-\delta t$ and $t$, respectively, and obtain feature correspondences using Lucas-Kanade optical flow method~\cite{Lucas81ijcai} (Fig.~\ref{fig:eye-catcher-four-images}(b)).
To cluster event features into groups that comply with different motion patterns (Fig.~\ref{fig:eye-catcher-four-images}(c)), we perform motion model fitting using feature correspondences with RANSAC and integrate it into the Progressive-X scheme~\cite{Barath2019iccv}.
In this scheme, motion model hypotheses are proposed one-by-one and added to a set of active instances maintained by the PEARL optimization~\cite{isack2012energy}, which looks for the optimal configuration over all instances and feature correspondences.
The result of this level consists of the labeling of features $\cL_\text{f}$ and associated motion instances $\cM$.
We detail the key components in this scheme as follows.\\
\newline
\noindent \textbf{Hypothesis Proposal.}
To propose sufficient motion model hypotheses, a proposal engine is needed to increasingly generate yet unseen hypotheses.
A variant of RANSAC algorithm -- Graph-Cut (GC) RANSAC~\cite{barath2018graph} is applied for this purpose.
The GC-RANSAC method consists of two key steps: sampling and inliers/outliers clustering.

1) Sampling. 
A minimal set $\cS$ is sampled, with which a motion model hypothesis is fitted using the linear solver (\ref{subsec:motion model fitting}).
Applicable sampler choices include NAPSAC~\cite{torr2002napsac} and PROSAC~\cite{chum2005matching}.
We utilize NAPSAC because of its local sampling characteristics. 

2) Inliers/outliers clustering. Inliers are separated from outliers through a binary classification process.
To this end, a neighbourhood graph is established, with each node representing a feature correspondence.
Actually, the topology of the neighbourhood graph is simply determined by the feature locations (nodes) on one IWE and a naive neighbour searching scheme determining the neighbourhood (edges), and thus, leading to a 2D graph (see zoom-in area of Fig. 1(c)).
The binary classification task is formulated as a standard Markov Random Field (MRF) problem, consisting of a data term and a smoothness term. 
The data term is designed in a way that rewards a data point close to a model if it is labeled as an inlier, while a modified \textit{Potts} model is applied as the smoothness term to deal with extreme cases in which the data contain significantly more outliers close to the desired model than inliers~\cite{barath2018graph}.  
The model hypothesis is then refitted with all inliers using a nonlinear optimizer.

In addition to these two steps, we need a quality function to verify a hypothesis.
The quality function's return reflects the support from data.
The design of the qualifying criteria is different from ordinary RANSAC algorithms specified for a single-instance fitting problem.
In a single-instance case, the quality function measures the goodness of fit w.r.t a unique hypothesis.
The simplest way is to count inliers.
On the contrary, in a multi-instance fitting problem, the quality function is supposed to count data that are inliers only to the hypothesis $h$, while being outliers to other existing instances $h_\cup$ (called compound model instance in~\cite{Barath2019iccv}).
In our problem, a data point $\mathbf{f}$ (a pair of corresponding features) is regarded as an inlier to a motion model hypothesis if the geometric error $e_{\mathbf{f}_i, \mathbf{m}_{\text{h}}} = \Vert \bx^{t}_i - W(\bx^{t-\delta t}_i; \mathbf{m}_{\text{h}})\Vert$ is smaller than $\zeta$ (a predefined threshold), 
and the minimal geometric error from other model instances $e_{\mathbf{f}_i, \mathbf{m}_{h_\cup}} = \min_{j \in \{h_\cup\}}\Vert \bx^{t}_i - W(\bx^{t-\delta t}_i; \mathbf{m}_j)\Vert$ is greater than $\zeta$. 
In practice, the truncated quality function of MSAC~\cite{torr2002bayesian} has shown its superiority to inlier counting of RANSAC in terms of accuracy and sensitivity to a predefined threshold.
Thus, we define our quality function $Q$ as
\begin{equation}
\label{eq:qualifying function}
Q(\mathbf{m}_h) = \vert \mathcal{F} \vert - \sum_{\mathbf{f} \in \mathcal{F}}\min \Big( 1, \max \Big( \frac{
e_{\mathbf{f}, \mathbf{m}_{\text{h}}}^{2}}{
\gamma(\zeta)^2}, 1-\frac{
e_{\mathbf{f}, \mathbf{m}_{h_\cup}}^{2}}{
\gamma(\zeta)^2}\Big) \Big),
\end{equation}
where $\vert \cdot \vert$ denotes the cardinality of a set, and $\gamma(\zeta) = \frac{3}{2}\zeta$.
As a consequence, a data point leads to a small score if it is consistent with both the newly proposed hypothesis and the existing compound instances or if it is inconsistent with the newly proposed hypothesis.

The above two steps are carried out iteratively until the probability of finding a hypothesis with better support is lower than a confidence threshold $\mu$.
As a consequence, the required number of iteration is calculated as $K = \frac{\log{(1-\mu)}}{\log{(1-\eta^{\vert \cS \vert})}}$,
where $\eta = \vert \mathcal{I} \vert / \vert \cF \vert$ denotes the inlier ratio.
The resulting hypothesis returned by the GC-RANSAC loop is added to the pool of active model instances, which participates in the following multi-instance optimization.\\

\noindent \textbf{Multi-Instance Optimization.}
The goal of this step, on one hand, is to find some of the data a better fit through iterative relabeling and re-fitting. 
On the other hand, the newly added hypothesis is further validated to see if it is redundant under the joint concern over model consistency and spatial coherence.
If not redundant, it will be maintained as an active instance in the model pool.
The formulation of this task is similar to the inliers/outliers classification mentioned above. However, a more complex multi-instance fitting problem is to be solved.
To this end, the aforementioned neighbourhood graph is re-used.
The PEARL~\cite{isack2012energy} optimization method is used to solve this problem.
Since the newly proposed hypotheses come one-by-one, the pool of instances is kept small, and thus, the computational complexity is not high.
Consequently, the pool of model instances is augmented, and the data are relabeled by refined model instances.\\
\newline

\setlength{\textfloatsep}{2pt}
\begin{algorithm}[t]
\SetAlgoLined
\SetKwInOut{Input}{input}
\SetKwInOut{Output}{output}
\Input{Feature correspondence set $\mathcal{F}$, threshold $\mu$}
\Output{Feature labeling $\mathcal{L}_\text{f}$, model instance $\mathcal{M}$} 
\text{Initialize} $\mathcal{L}_\text{f} = \varnothing$, $\mathcal{M} = \varnothing$, $\omega^{\star} = 0$, $\bar{I} = \vert \cS \vert + 2$\\
$\mathcal{G} \leftarrow \text{Neighbourhood graph } (\mathcal{F})$\\
\While(\Comment*[f]{\text{Prog-X Loop}~\cite{Barath2019iccv}}){$\bar{I} > \vert \cS \vert + 1$}
{
\For(\Comment*[f]{\text{GC-RANSAC Loop}~\cite{barath2018graph}}){$k = 1 \text{ to } K$}
{
$\mathcal{S}$ $\leftarrow$ $\text{Sampled by NAPSAC}$ \cite{torr2002napsac}\\
$\mathbf{m}_h \leftarrow \text{Linear model fitting } (\mathcal{S})$\\
$\omega \leftarrow \text{Compute support of } \mathbf{m}_h $ (Eq.~\ref{eq:qualifying function})\\
\While{$\omega > \omega^{\star}$}{
$\omega^{\star} \leftarrow \omega$, $\mathbf{m}_h^{\star} \leftarrow \mathbf{m}_h$, $\mathcal{I}^{\star} \leftarrow \mathcal{I}$\\
$\mathcal{I} \leftarrow  \text{Graph-Cut } (\mathbf{m}_h, \mathcal{G})$\\
$\mathbf{m}_h \leftarrow \text{Nonlinear model fitting }(\mathcal{I}$)\\
$\omega \leftarrow \text{Compute support of } \mathbf{m}_h $ (Eq.~\ref{eq:qualifying function})\\
}
}
$\cM \leftarrow \cM \oplus \mathbf{m}_h^{\star}$\\
$\{\mathcal{L}_\text{f}, \mathcal{M}\} \leftarrow \text{PEARL} (\cM, \mathcal{F})$~\cite{isack2012energy}\\
$\bar{I} \leftarrow \text{Compute the upper limit of } \vert \mathcal{I}^{\star} \vert $ (Eq.~\ref{eq:prog-x termination criteria})
}
\caption{Feature Clustering with Progressive Multi-Model Fitting
}
\label{alg: event clustering with prog-x}

\end{algorithm}

\noindent \textbf{Termination Criterion}
The progressive multi-model fitting pipeline, summarized in Alg.~\ref{alg: event clustering with prog-x}, runs iteratively until the termination criterion is met.
According to~\cite{Barath2019iccv}, the overall progress is terminated when $\bar{I} < \vert \mathcal{S} \vert + 1$, where $\bar{I}$ denotes the upper bound of the number of inliers to a not yet found instance.
The inliers in this context refer to data that are independent on the compound model.
The upper bound $\bar{I}$ with confidence $\mu$ in the $n_\text{th}$ iteration is defined as
\begin{equation}
\label{eq:prog-x termination criteria}
\bar{I}(\cF, h_{\cup}, \vert \cS \vert, n, \mu) = (\vert \cF \vert - \vert h_{\cup} \vert)\sqrt[\vert \cS \vert]{1-\sqrt[n]{1-\mu}},
\end{equation}
where, with a slight abuse of notation, $\vert h_{\cup} \vert$ denotes the number of inliers to the compound model.
This criterion guarantees that, upon termination, the probability of an existing unseen instance with at least $\vert \cS \vert + 1$ inliers is smaller than $1-\mu$.

\subsection{Event Clustering via Spatio-Temporal Graph Cut}
\label{subsec: level two}

The goal of this step is to achieve event-wise labeling such that each event is associated with a specific motion model.
We borrow the idea of event-based motion segmentation in~\cite{zhou2020eventbased}, which also formulates the segmentation problem as an energy-based multi-model fitting problem.
In contrast to the 2D neighbourhood graph applied in \ref{subsec: level one}, a 3-D fully connected graph is established by exploiting the spatio-temporal characteristics of events.
The proposed objective is a hybrid function of discrete labeling $L$ and continuous model parameters $\cM$, defined as
\begin{equation}
E(L, \cM) \doteq E_{\text{D}}(L, \cM) + E_{\text{P}}(L) + E_{\text{M}}(L),
\label{eq: objective level 2}
\end{equation}
which simultaneously considers the data-model consistency (data term $E_{\text{D}}$) and spatio-temporal coherence across the data (\textit{Potts} model $E_{\text{P}}$).
The data term considers the goodness of fit by adapting the idea of motion compensation, while the smoothness term enhances spatially coherent labeling.
Additionally, an \textit{Minimum Description Length} (MDL) term ($E_{\text{M}}$) is added in order to make the model pool as compact as possible.
This constraint would reduce model redundancy and thus prompt the remaining models to correspond exactly to IMOs in the scene.
Due to space limitations, the definition on each term is omitted.
Readers can refer to \cite{zhou2020eventbased} for more details.
A block-coordinate descent strategy is used to solve Eq.~\ref{eq: objective level 2}, optimizing $L$ and $\cM$ in an alternating manner.
\begin{figure}[t]
\vspace{-1cm}
\end{figure}
The original initialization method of \cite{zhou2020eventbased} proposes hypotheses by fitting models to events within a number of hierarchically divided spatio-temporal volumes.
This strategy, though free of optical flow computation, generates many similar (redundant) hypotheses together with false-positive ones, and thus, is relatively slow.
Besides, it typically fails to detect small IMOs because of the bad inlier ratio.
Thanks to the feature clustering in the first level, the resulting model instances can be used as the initial pool of models.
With this compact and reasonably accurate pool of motion instances, the inference of event-model association can be done efficiently.
We evaluate our algorithm in the following section.

\section{Experiments}
\label{sec:evaluation}

In this section, we evaluate the proposed algorithm.
First, we introduce the datasets and evaluation metrics used (Section~\ref{sec:experiment:dataset}).
Second, we provide quantitative and qualitative evaluation results on these datasets (Section \ref{sec:experiment:quantitative} and \ref{sec:experiment:qualitative}).
Then we discuss the parameters used and the computational performance (Section \ref{sec:experiment:computational}).
Finally, we conclude with the limitations of the proposed method (Section \ref{sec:experiment:limitation}).

\begin{table}[b]
\vspace{0.4cm}
\caption{Summary of datasets' characteristics}
\label{table: dataset summary}
\centering
\begin{adjustbox}{max width=\columnwidth}
\setlength{\tabcolsep}{2pt}
\begin{tabular}{c|ccccccl}
\toprule

Dataset & Camera & Camera status& \#IMO & Env. & HDR & Non-rigid\\
\midrule
EED \cite{Mitrokhin18iros}& DAVIS240 &Moving & 1-3 & Indoor & Yes & No\\
EV-IMO \cite{Mitrokhin19iros} & DAVIS346&Moving & 1-3 & Indoor & No & No\\
DistSurf \cite{almatrafi2020distance}& DAVIS240& Static & 1-3 & In/Outdoor & No & No\\
EMSMC \cite{Stoffregen19iccv} & DAVIS346 &Moving& 1-3 & In/Outdoor & Yes & Yes\\
Our Data & DAVIS346&Moving & 1-3 & Indoor & No & Yes\\

\bottomrule %
\end{tabular}

\end{adjustbox}
\end{table}
\subsection{Datasets and Evaluation Metrics}
\label{sec:experiment:dataset}

To extensively evaluate the algorithm, we use almost all publicly available datasets~\cite{Mitrokhin18iros, Mitrokhin19iros, Stoffregen19iccv, almatrafi2020distance}.
Except for \textit{DistSurf}~\cite{almatrafi2020distance} using a static event camera, all the others were captured with a hand-held moving camera, in either indoor or outdoor environments.
The number of IMOs was up to three, and challenging cases including HDR scenes and non-rigid motions were also considered.
Details of these datasets are summarized in Table.~\ref{table: dataset summary}.
Two standard metrics are used in the quantitative evaluation, including \textit{detection rate} and \textit{Intersection over Union (IoU)}.
Details about the metrics can be found in \cite{Mitrokhin18iros, zhou2020eventbased}.

\begin{table}[t]
\vspace{0.2cm}
\caption{Quantitative evaluation on EED dataset~\cite{Mitrokhin18iros} using detection rate of IMOs (in \%).}
\label{table: eed}
\centering
\begin{adjustbox}{max width=\columnwidth}
\begin{threeparttable}
\begin{tabular}{lccccc}
\toprule
EED sequence & EED\cite{Mitrokhin18iros}&EMSMC\cite{Stoffregen19iccv} &0-MMS \cite{parameshwara20200mms} & EMSGC \cite{zhou2020eventbased} &Ours \\
\midrule
what is background&89.21&\textbf{100.0} & -\tnote{*} & \textbf{100.0} & \textbf{100.0} \\
occlusion & 90.83&92.31 &- & \textbf{100.0}&\textbf{100.0} \\
fast drone & 92.78 &\textbf{96.30}& - & \textbf{96.30} &\textbf{96.30}\\
light variation &84.52 & 80.51 & - & \textbf{93.51}&92.21 \\
multiple objects &87.32 &\textbf{96.77} & - & 0 & 95.67 \\
\midrule
Average & 88.93 & 93.18 & 94.2 & 77.96 & \textbf{96.84} \\
\bottomrule %
\end{tabular}

\begin{tablenotes}
        \footnotesize
        \item[*] Numbers are not available for each individual sequence in~\cite{parameshwara20200mms}.
      \end{tablenotes}
    \end{threeparttable}
    
\end{adjustbox}
\end{table}
\begin{table}[t]
\caption{Quantitative evaluation on EV-IMO dataset~\cite{Mitrokhin19iros} using the IoU metric (in\%)}
\label{table: evimo}
\centering
\begin{adjustbox}{max width=\columnwidth}
\setlength{\tabcolsep}{2pt}
\begin{tabular}{c|ccccl}
\toprule
Method & EV-IMO\cite{Mitrokhin19iros} & MOMS\cite{parameshwara2020moms} & 0-MMS \cite{parameshwara20200mms}& EMSGC \cite{zhou2020eventbased} & Ours\\
\midrule
IoU & 77.00 & 74.82 & 80.37 &76.81 & \textbf{80.73}\\
\bottomrule %
\end{tabular}

\end{adjustbox}
\vspace{-0.6cm}
\end{table}

\subsection{Quantitative Evaluation}
\label{sec:experiment:quantitative}

\textbf{EED Dataset}~\cite{Mitrokhin18iros}. 
We evaluate our algorithm on all the five sequences using the \textit{detection rate} metric and compared against state-of-the-art approaches~\cite{Mitrokhin18iros, Stoffregen19iccv, parameshwara20200mms, zhou2020eventbased}.
As shown in Table~\ref{table: eed}, our algorithm overall outperforms the other methods.
Note that our method gave less accurate result than EMSGC~\cite{zhou2020eventbased} on the \textit{light variation} sequence.
The problem was caused by false-positive feature correspondences used as input in the first level.
These features were extracted from noisy events induced by a strobe light flashing in the background.
Such a bad information-noise ratio generated many false-positive motion model hypotheses, leading to inaccurate segmentation results in the second level.
Nevertheless, EMSGC~\cite{zhou2020eventbased} failed to detect small-size IMOs (\textit{Multiple objects} sequence) because of its initialization that struggled in the case of bad inlier-outlier ratio, whereas our method can handle this due to the proposed feature tracking.

\global\long\def\figWidth{0.22\columnwidth}
\begin{figure}[th]
\vspace{0.2cm}
	\centering
	\quad\, %
    {\small
    \setlength{\tabcolsep}{2pt}
	\begin{tabular}{
	>{\centering\arraybackslash}m{0.5cm} 
	>{\centering\arraybackslash}m{\figWidth} 
	>{\centering\arraybackslash}m{\figWidth} 
	>{\centering\arraybackslash}m{\figWidth}
	>{\centering\arraybackslash}m{\figWidth}
	>{\centering\arraybackslash}m{\figWidth}}
    
    \rotatebox{90}{\makecell{Background}}
    &\frame{\includegraphics[width=\linewidth]{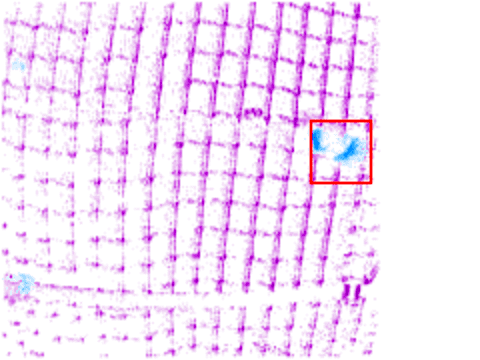}}
    & \frame{\includegraphics[width=\linewidth]{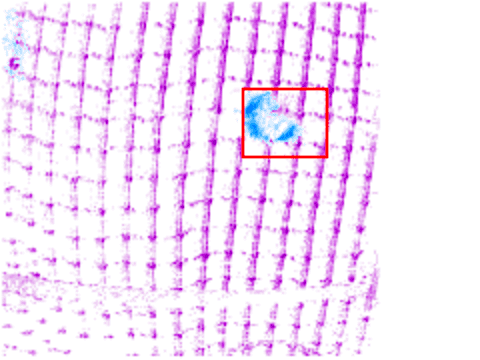}}
    & \frame{\includegraphics[width=\linewidth]{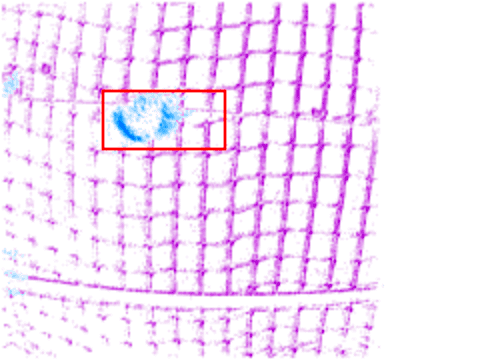}}
    &\frame{\includegraphics[width=\linewidth]{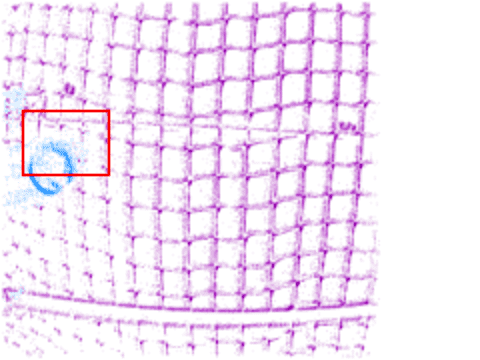}}
    \\
    
    \rotatebox{90}{\makecell{Occlusion}}
    &\frame{\includegraphics[width=\linewidth]{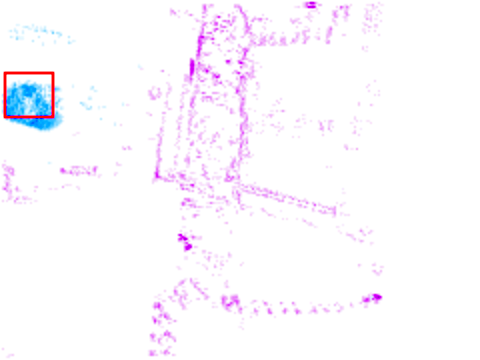}}
     &\frame{\includegraphics[width=\linewidth]{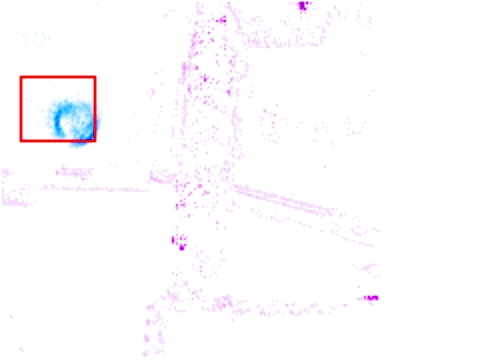}}
    &\frame{\includegraphics[width=\linewidth]{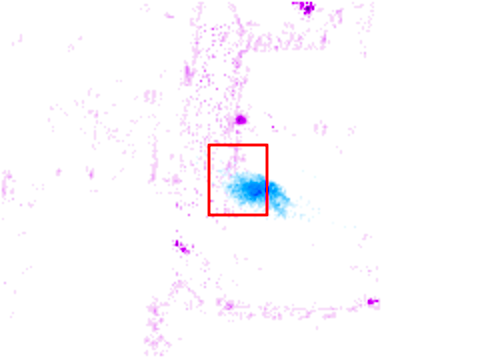}}
     &\frame{\includegraphics[width=\linewidth]{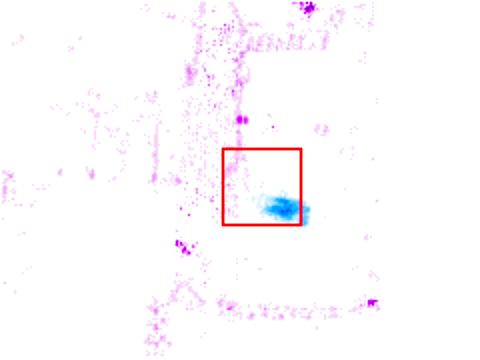}}
    \\
    
    \rotatebox{90}{\makecell{Fast \\ drone}}
    &\frame{\includegraphics[width=\linewidth]{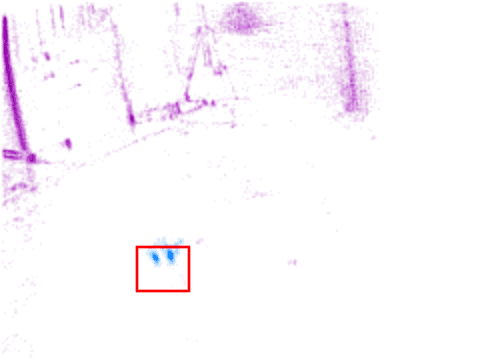}}
   & \frame{\includegraphics[width=\linewidth]{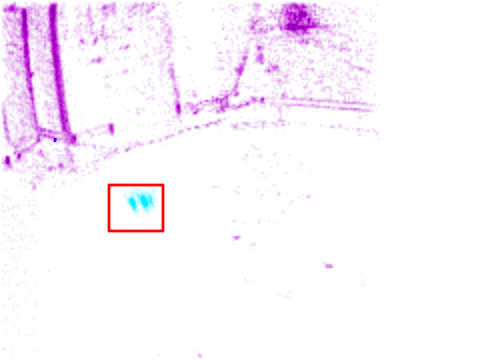}}
    &\frame{\includegraphics[width=\linewidth]{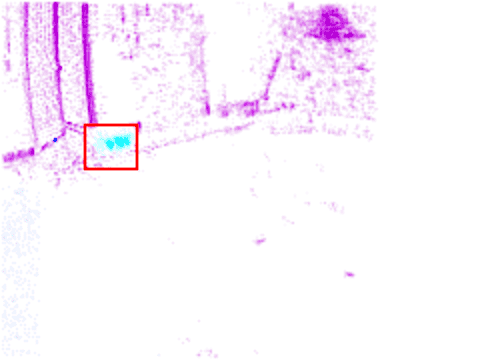}}
    &\frame{\includegraphics[width=\linewidth]{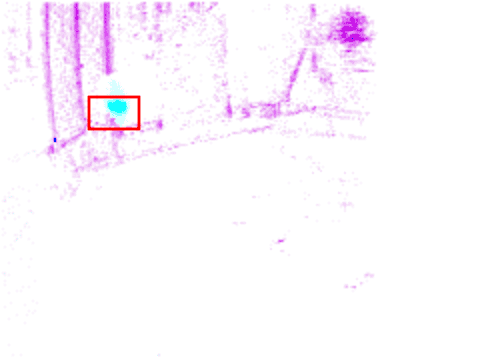}}
    \\
    
    \rotatebox{90}{\makecell{Light \\ variation}}
    &\frame{\includegraphics[width=\linewidth]{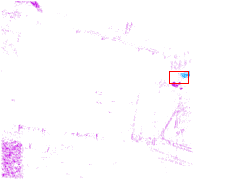}}
    &\frame{\includegraphics[width=\linewidth]{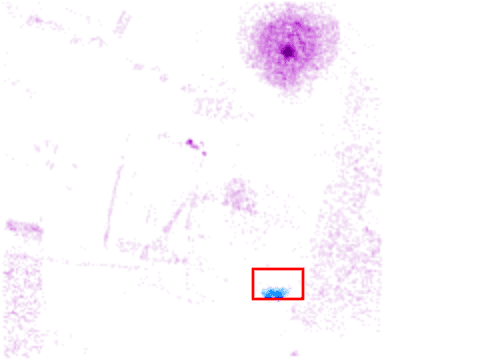}}
    &\frame{\includegraphics[width=\linewidth]{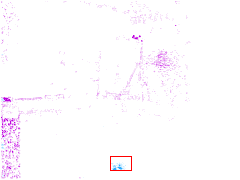}}
     &\frame{\includegraphics[width=\linewidth]{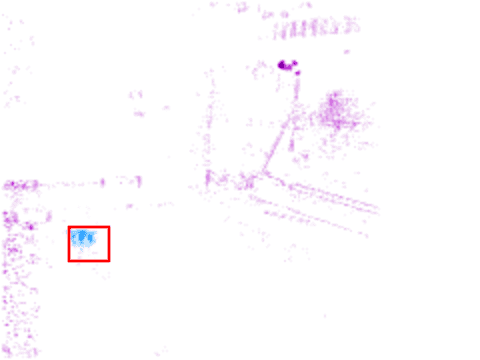}}
    \\
    
    \rotatebox{90}{\makecell{Multiple \\ objects}}
    &\frame{\includegraphics[width=\linewidth]{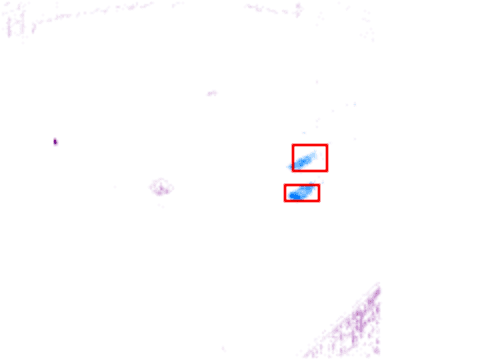}}
    &\frame{\includegraphics[width=\linewidth]{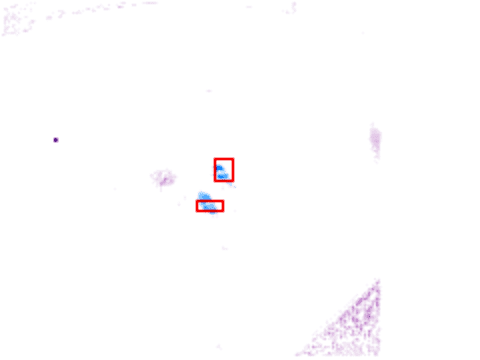}}
    &\frame{\includegraphics[width=\linewidth]{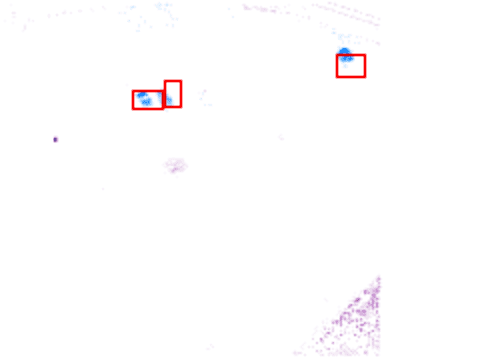}}
    &\frame{\includegraphics[width=\linewidth]{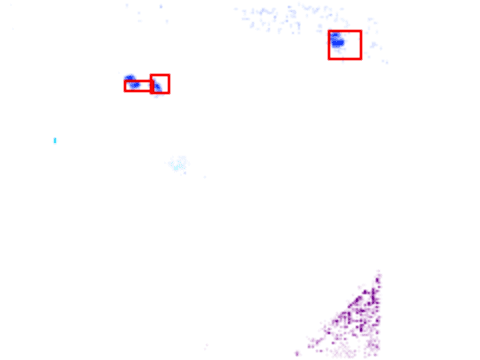}}
    \\

	\end{tabular}
	}
	\caption{Segmentation result on EED dataset~\cite{Mitrokhin18iros}.
    The red bounding boxes show the location of the IMOs.
    Note that all ground truth bounding boxes are manually annotated on the DAVIS grayscale images that are not temporally aligned with event data.
    Hence, offsets are witnessed for fast moving objects.
    }
	\label{fig: eed sequence}
\vspace{-0.6cm} 
\end{figure}

\textbf{EV-IMO Dataset}~\cite{Mitrokhin19iros}. 
We also evaluate our algorithm using the \textit{IoU} metric when ground-truth dense segmentation masks are provided.
As can be seen from the quantitative results in Table~\ref{table: evimo}, our algorithm outperforms the other state-of-the-art solutions.
The numbers for baseline methods are still taken from the corresponding publications.
Qualitative results are given in Fig~\ref{fig: evimo sequence}, where we respectively show the labelled features, the labelled events (IWEs), and the dense segmentation masks (convex hulls) overlapped on corresponding grayscale images.
The results demonstrate that objects can be successfully separated from the background as long as they undergo motions independent of the camera's ego-motion.
Note that the feature labeling result from level one is sometimes not perfect. 
We observed redundant motion model instances, and some weak features from the same IMO that may be associated to different motion models.
Both of these two issues can be resolved in level two thanks to the applied regularization terms (\textit{Potts} model and MDL term), as long as true-positive motion model proposals are obtained from level one.

\global\long\def\figWidth{0.15\linewidth}
\begin{figure*}[t]
\vspace{0.2cm}
	\centering
	\quad\, %
    {\small
    \setlength{\tabcolsep}{2pt}
	\begin{tabular}{
	>{\centering\arraybackslash}m{\figWidth} 
	>{\centering\arraybackslash}m{\figWidth}
	>{\centering\arraybackslash}m{\figWidth}
	>{\centering\arraybackslash}m{\figWidth} 
	>{\centering\arraybackslash}m{\figWidth}
	>{\centering\arraybackslash}m{\figWidth}}

        {} & 
        Box\cite{Mitrokhin19iros} &
        {} & 
        {} & 
        Table \cite{Mitrokhin19iros} &
        {}
		\\

        \\
				\frame{\includegraphics[width=\linewidth]{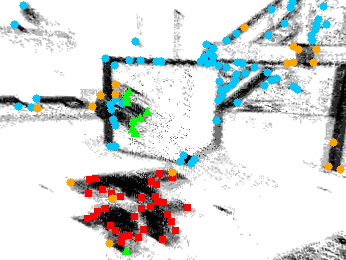}}
				&\frame{\includegraphics[width=\linewidth]{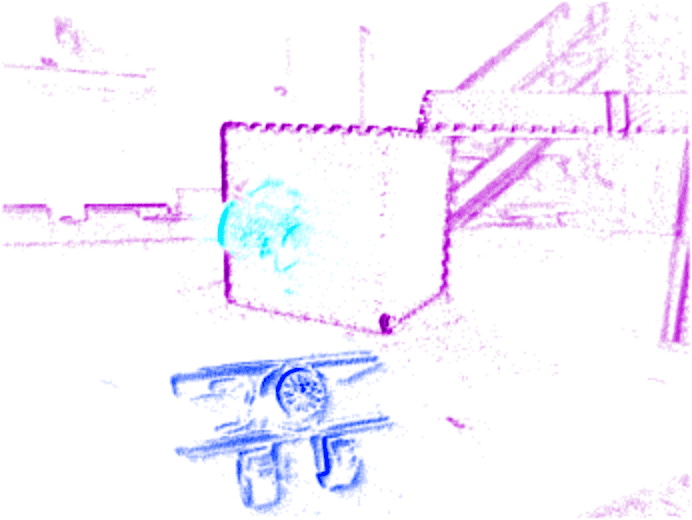}}
                &\frame{\includegraphics[width=\linewidth]{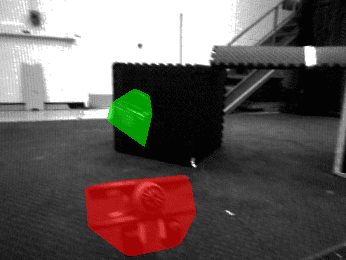}}
                &\frame{\includegraphics[width=\linewidth]{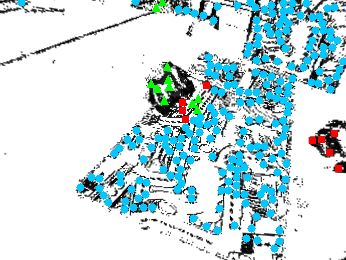}}
                 &\frame{\includegraphics[width=\linewidth]{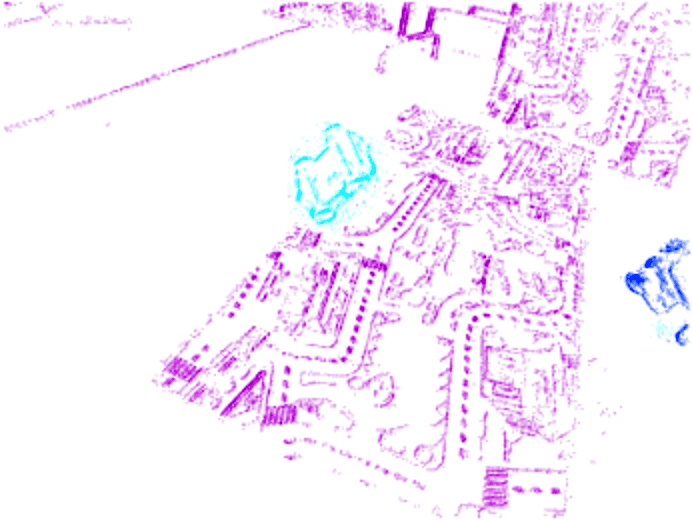}}
                 &\frame{\includegraphics[width=\linewidth]{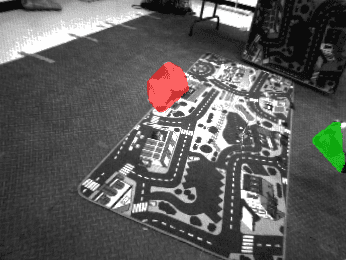}}
        \\
        
      			\frame{\includegraphics[width=\linewidth]{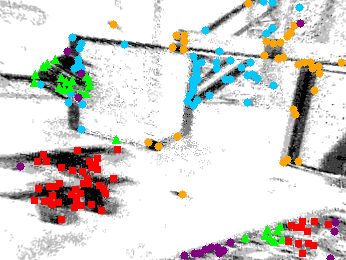}}
        		&\frame{\includegraphics[width=\linewidth]{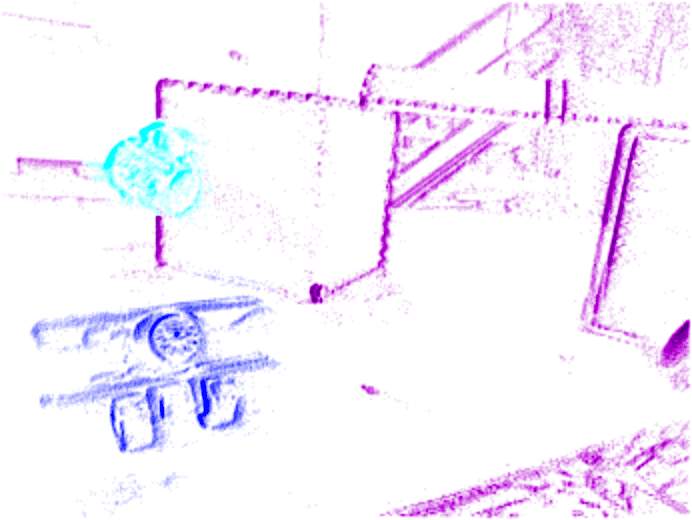}}
                &\frame{\includegraphics[width=\linewidth]{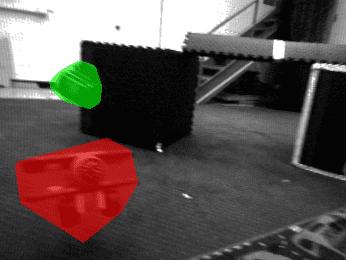}}
                &\frame{\includegraphics[width=\linewidth]{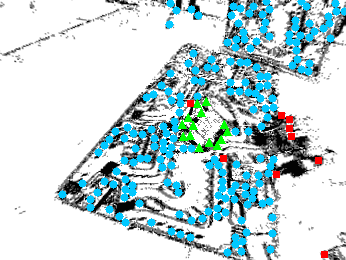}}
                 &\frame{\includegraphics[width=\linewidth]{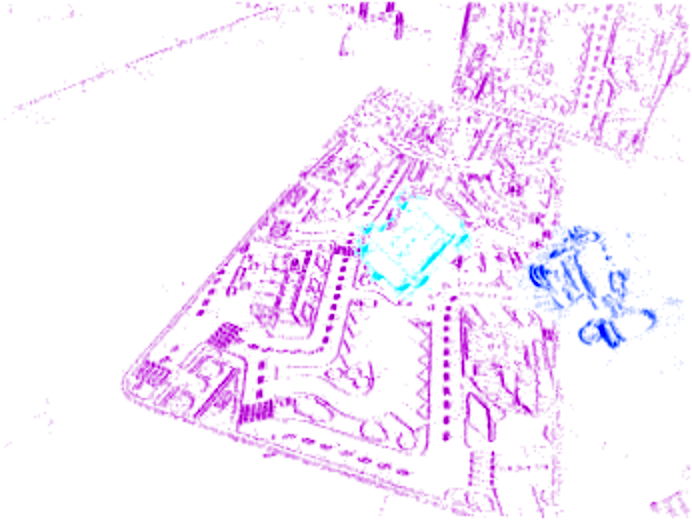}}
                 &\frame{\includegraphics[width=\linewidth]{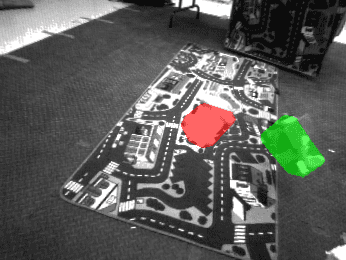}}
        \\
        		\frame{\includegraphics[width=\linewidth]{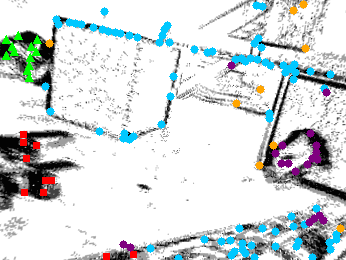}}
        		&\frame{\includegraphics[width=\linewidth]{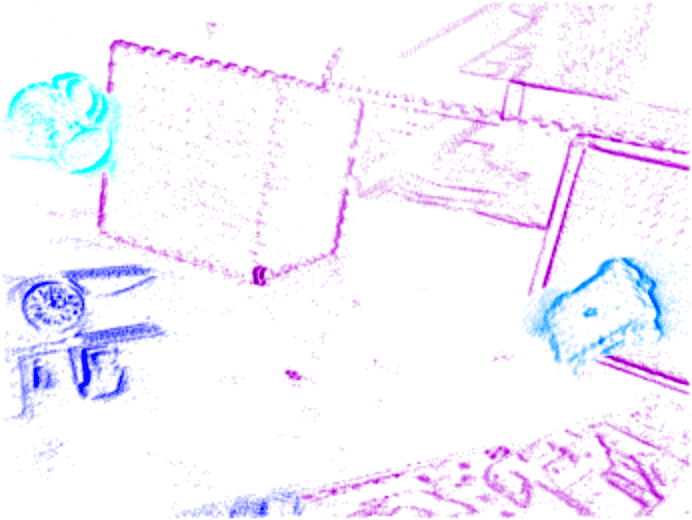}}
                &\frame{\includegraphics[width=\linewidth]{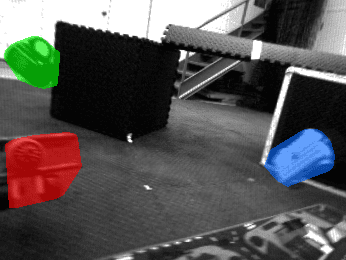}}
                &\frame{\includegraphics[width=\linewidth]{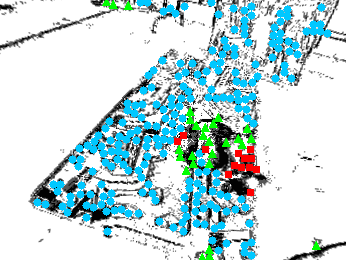}}
                &\frame{\includegraphics[width=\linewidth]{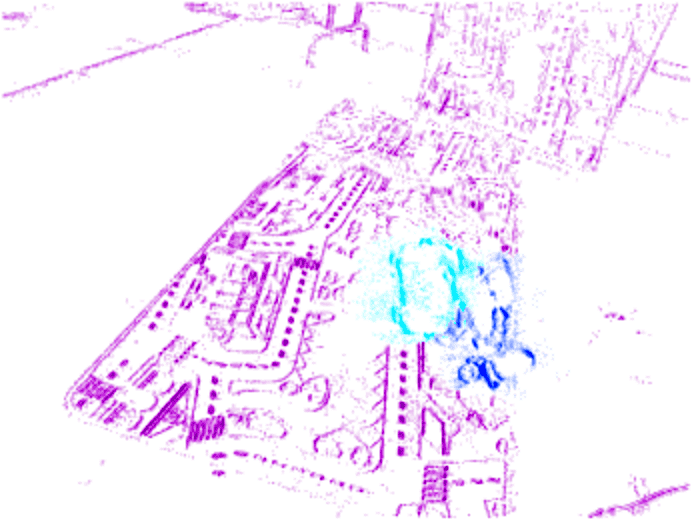}}
                 &\frame{\includegraphics[width=\linewidth]{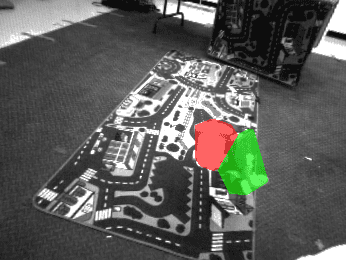}}
        \\

	\end{tabular}
	}
	\caption{Segmentation results on the EV-IMO dataset~\cite{Mitrokhin19iros}. 
	Column 1 and 4 show the feature labeling result %
	of two sequences: \textit{Box} and \textit{Table}.
	Features clustered to the same group are labeled with an identical color.
	Column 2 and 5 show the labeled IWEs.
	Note that we use a different color scheme for the visualization of IWEs than that for feature labeling.
	Column 3 and 6 show the corresponding dense segmentation masks on the corresponding grayscale images from DAVIS~\cite{Brandli14ssc}. 
	The graysclae images are for visualization only.}

	\label{fig: evimo sequence}
	\vspace{-2ex}
\end{figure*}

\subsection{Qualitative Evaluation}
\label{sec:experiment:qualitative}
In addition to the above quantitative results, we also provide extensive qualitative evaluation on a set of real-world data, including DistSurf dataset~\cite{almatrafi2020distance}, EMSMC dataset~\cite{Stoffregen19iccv}, and our data. 
Exemplary results are shown in Fig.~\ref{fig: further real world sequence}.

The DistSurf dataset consists of two sequences: cars and hands.
As shown in the first two columns of Fig.~\ref{fig: further real world sequence}, 
our algorithm successfully separates an IMO from the other, even when they overlap.
Results of the two selected sequences from EMSMC dataset are illustrated in the third and fourth column of Fig.~\ref{fig: further real world sequence}.
The \textit{skateboarder} sequence was captured in an outdoor HDR scene with the event camera looking towards the sun while a pedestrian and a skateboarder pass by.
This result demonstrates that event-based solutions are qualified in HDR scenarios.
Also, the compact segmentation is due to the MDL term applied in level two, which approximates complex non-rigid motions with a small number of rigid ones.
The \textit{fan \& coin} sequence captures a rotating fan (rotation speed around 1800 rpm) and a free-falling coin.
Our algorithm supports different types of motion models, and thus allows us to model and separate the rotating fan blades and the falling coin.
Our data captured indoor scenes with multiple people passing by.
It is more complicated compared to the aforementioned ones because multiple non-rigid motions and occlusions coexist. 
Our algorithm performs well on these sequences.
\global\long\def\figWidth{0.15\linewidth}
\begin{figure*}[t]
\vspace{0.2cm}
	\centering
	\quad\, %
    {\small
    \setlength{\tabcolsep}{2pt}
	\begin{tabular}{
	>{\centering\arraybackslash}m{\figWidth} 
	>{\centering\arraybackslash}m{\figWidth} 
	>{\centering\arraybackslash}m{\figWidth} 
	>{\centering\arraybackslash}m{\figWidth}
	>{\centering\arraybackslash}m{\figWidth}
	>{\centering\arraybackslash}m{\figWidth}}
			
		Cars~\cite{almatrafi2020distance} & Hands~\cite{almatrafi2020distance} & Skateboarder~\cite{Stoffregen19iccv} & Fan\&Coin~\cite{Stoffregen19iccv} & 
		Downstairs (ours) &
		Indoor (ours)\\

		\frame{\includegraphics[width=\linewidth]{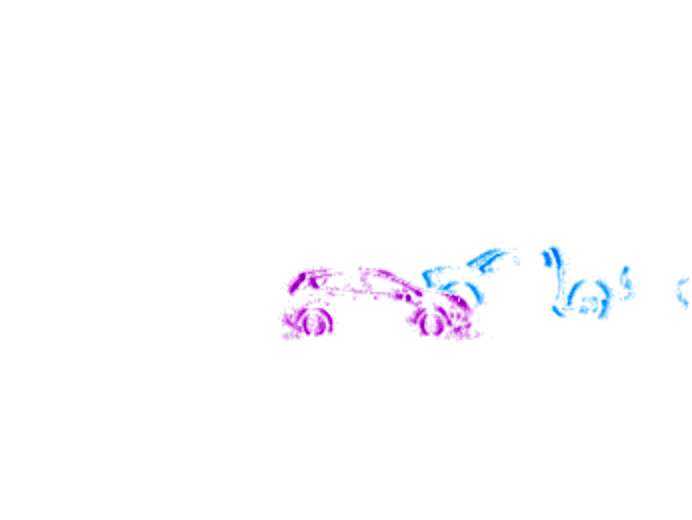}}
		&\frame{\includegraphics[width=\linewidth]{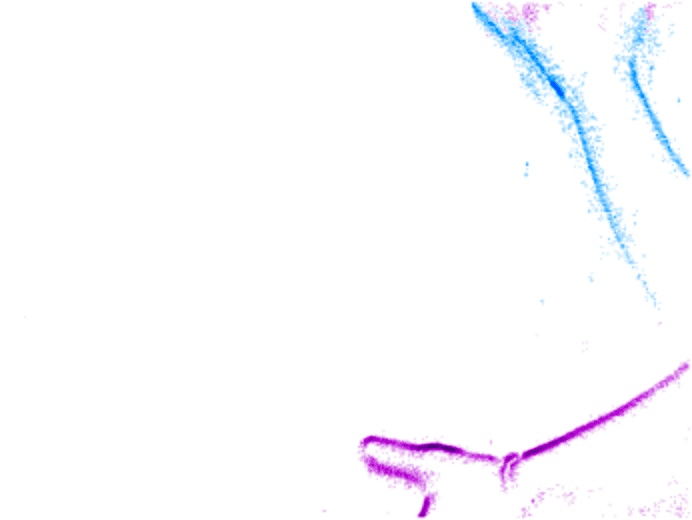}}
		&\frame{\includegraphics[width=\linewidth]{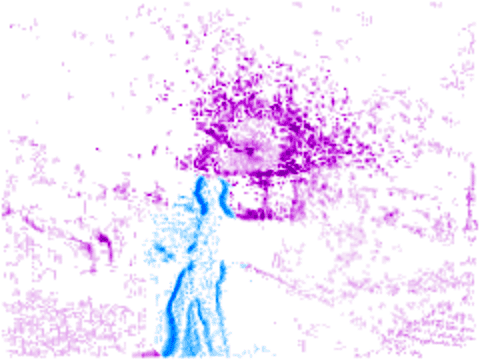}}
		&\frame{\includegraphics[width=\linewidth]{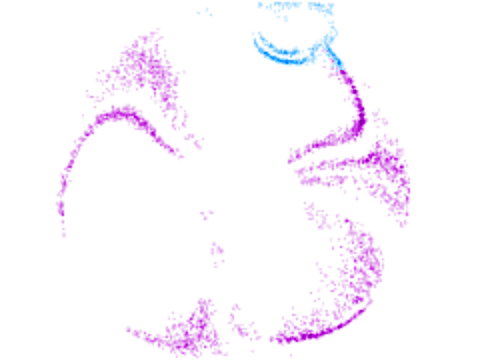}}
		&\frame{\includegraphics[width=\linewidth]{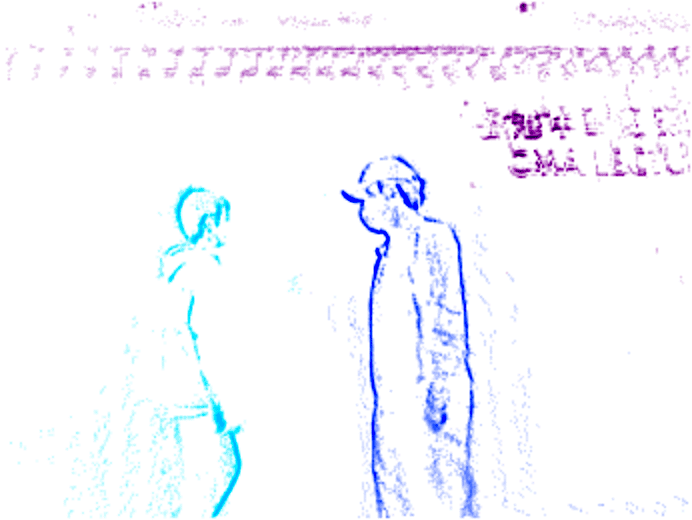}}
		&\frame{\includegraphics[width=\linewidth]{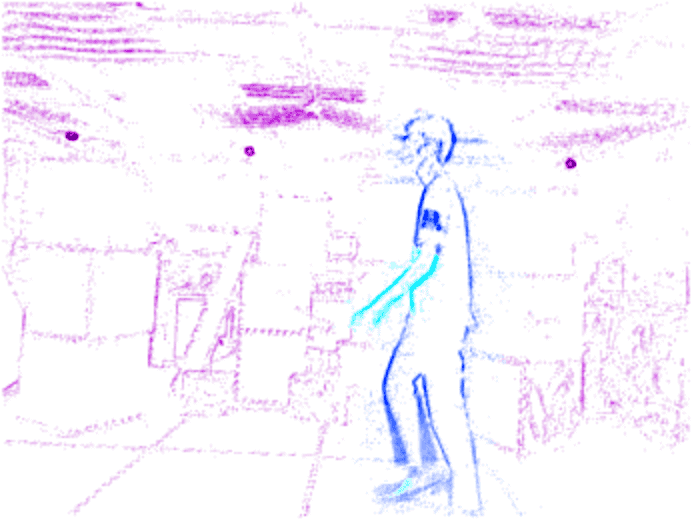}}
		\\
		
		\frame{\includegraphics[width=\linewidth]{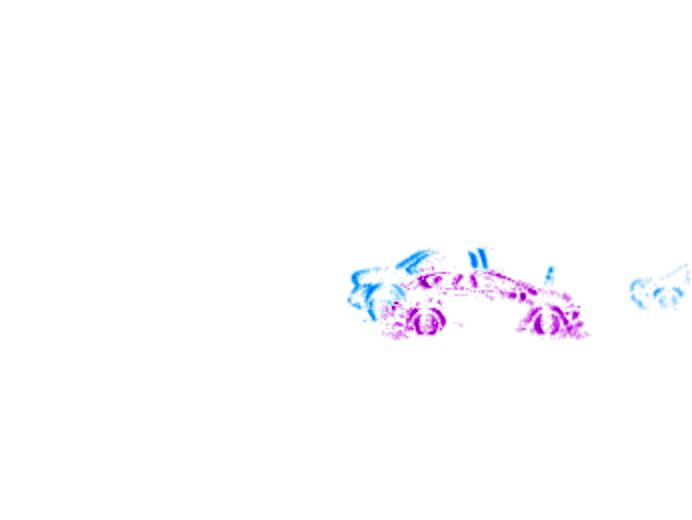}}
		&\frame{\includegraphics[width=\linewidth]{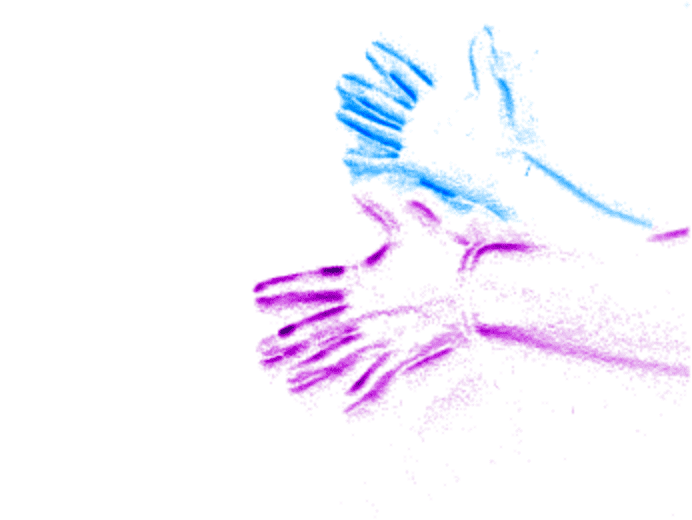}}
		&\frame{\includegraphics[width=\linewidth]{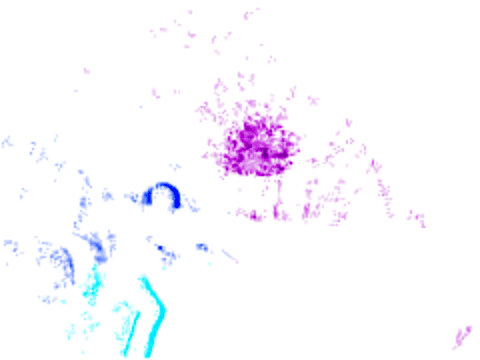}}
		&\frame{\includegraphics[width=\linewidth]{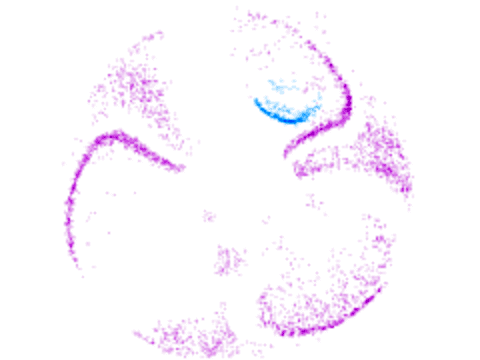}}
		&\frame{\includegraphics[width=\linewidth]{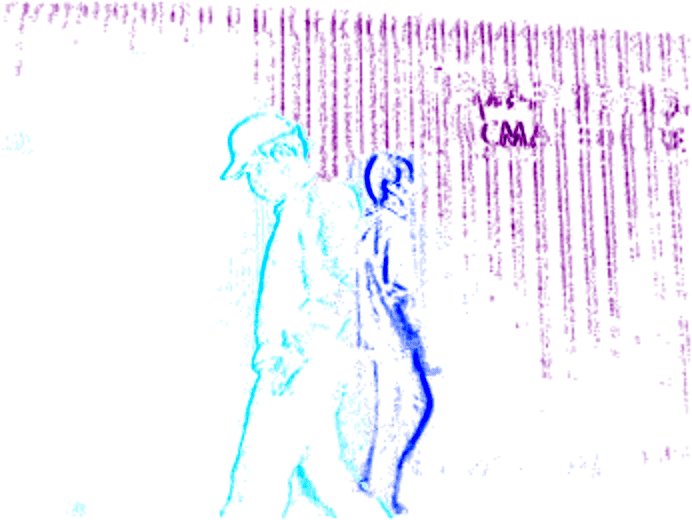}}
		&\frame{\includegraphics[width=\linewidth]{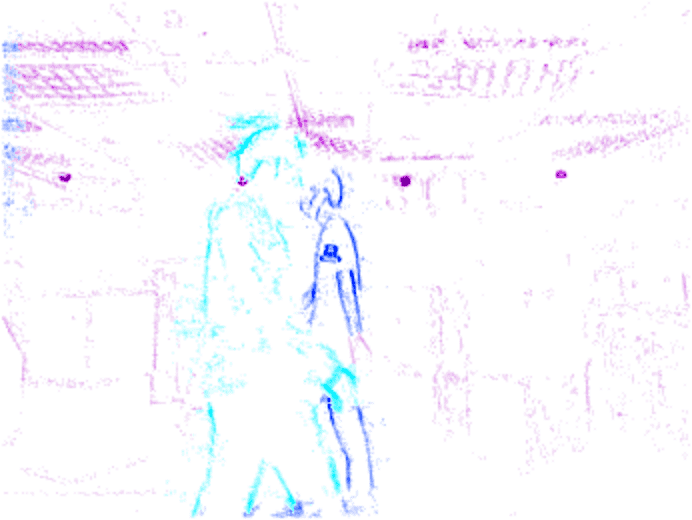}}
		\\
		
		\frame{\includegraphics[width=\linewidth]{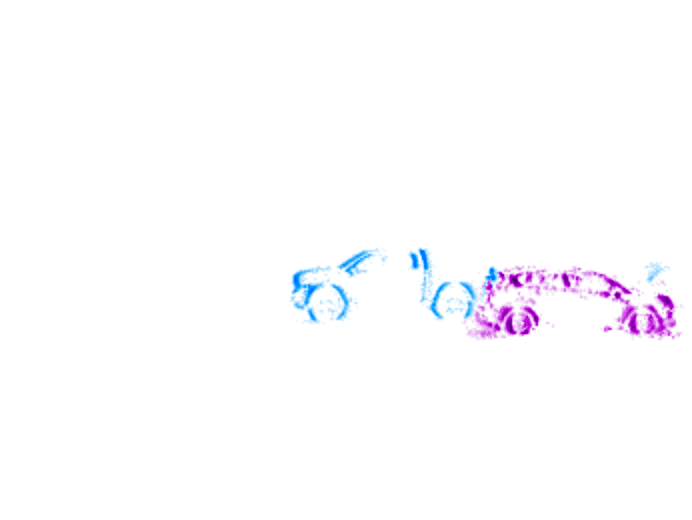}}
		&\frame{\includegraphics[width=\linewidth]{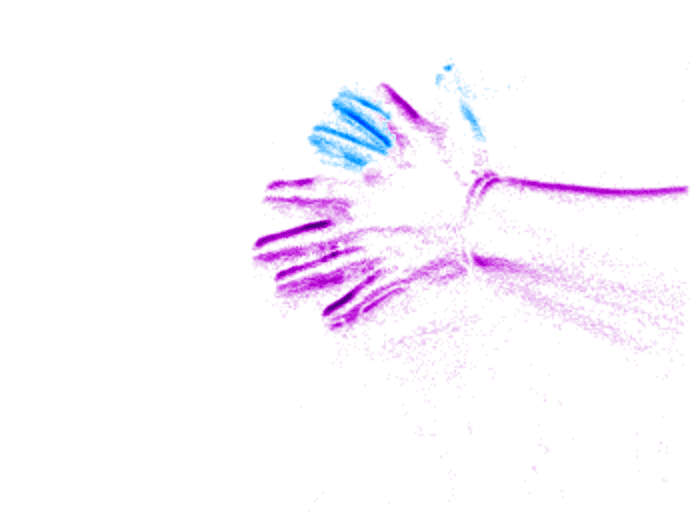}}
		&\frame{\includegraphics[width=\linewidth]{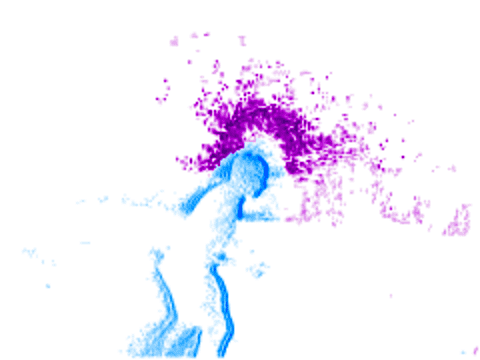}}
		&\frame{\includegraphics[width=\linewidth]{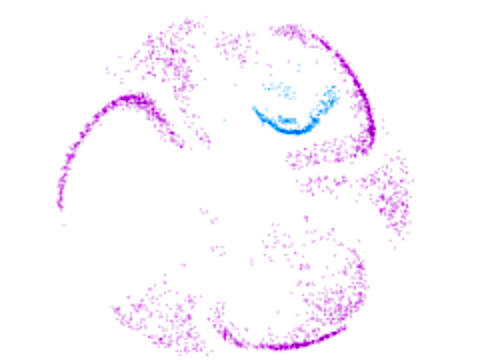}}
		&\frame{\includegraphics[width=\linewidth]{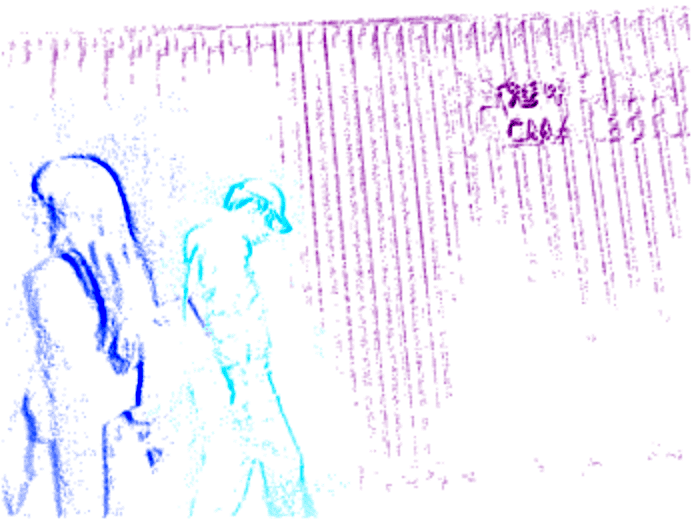}}
		&\frame{\includegraphics[width=\linewidth]{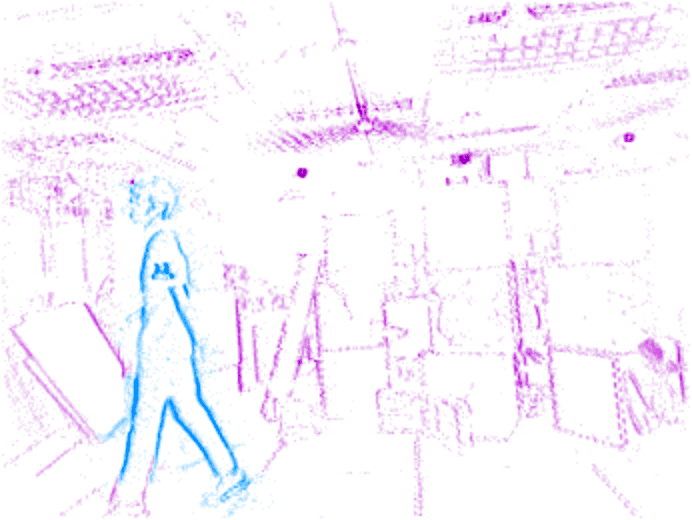}}
		\\

	\end{tabular}
	}
	\caption{Qualitative results on a set of real-world data. }
	\label{fig: further real world sequence}
	\vspace{-0.5ex}
\end{figure*}
\subsection{Parameters and Computational Performance}
\label{sec:experiment:computational}

The segmentation is performed on events occurred within a constant time interval, and we process events in a fashion of sliding window.
The width of the temporal window $\delta t$ is set by $15 \textendash 35$ ms according to the scene dynamics and textures.
The confidence threshold $\mu$ is set to $0.95$.
The inlier-outlier (geometric error in Section~\ref{subsec: level one}) threshold is set to $1.5$ pixel.
The parameters used at level two are taken from~\cite{zhou2020eventbased}.

We ran our algorithm using a single thread on an Intel i7-8700K CPU.
The average computation time of each module in Fig.~\ref{fig:flowchart} is listed in Table~\ref{table: computation time}.
It is worth mentioning that the feature clustering algorithm at level one performs thousands of iterations to propose hypotheses extensively.
To speed up, we reuse historical motion model to bootstrap the estimation at next time, and thus, only data corresponding to new appearing IMOs need to be repeatedly sampled and validated.
Numbers in parentheses refer to the time when reusing historical models.
With the resulting compact pool of models, the spatio-temporal graph cut method at level two can be finished efficiently. 
The overall computation time for each independent segmentation takes 0.36 second approximately, which is 20 times faster than~\cite{zhou2020eventbased}.

\begin{table}[b]
\vspace{0.3cm} 
\caption{Computation time of our algorithm}
\label{table: computation time}
\centering
\begin{adjustbox}{max width=\columnwidth}
\setlength{\tabcolsep}{2pt}
\begin{tabular}{c|lccc}
\toprule
Module         & \begin{tabular}[c]{@{}l@{}}Feature Detection\\ and Tracking\end{tabular} & \begin{tabular}[c]{@{}c@{}}Feature Clustering\\ (Level One)\end{tabular} & \begin{tabular}[c]{@{}c@{}}Event Clustering\\ (Level Two)\end{tabular} & Overall \\
\midrule
Avg. Time (ms) & \multicolumn{1}{c}{3} & 110 (15) & 250 & 363 (268)\\
\bottomrule %
\end{tabular}
\end{adjustbox}
\end{table}

\subsection{Limitations}
\label{sec:experiment:limitation}

Two limitations of the proposed method are witnessed during our experiment.
First, our method cannot distinguish two IMOs under similar motion.
For example, two cars moving from right to left in the \textit{Cars} sequence \cite{almatrafi2020distance} are likely to be recognized as one IMO no matter how far they are from each other.
This issue can be resolved by introducing semantic cues.
Second, the width of the temporal window has to be selected according to the characteristics of each data.
This is due to the fact that the applied linear motion model is held only within a short time interval, thus a narrow window is preferred.
On the other hand, the applied motion-compensation scheme needs enough textures to evaluate the sharpness, thus requiring a relatively wide window.
The parameter $\delta t$ is set by making a trade-off between the two considerations.
A better way to resolve this limitation is to apply nonlinear motion models, such as a B\'{e}zier spline \cite{seok2020robust}, which is able to describe complex motion within a relatively longer temporal window.
We leave these to future work.
\section{Conclusion}
\label{sec:conclusion}

We present a novel approach for event-based motion segmentation in this paper.
Our method is a cascaded two-level multi-model fitting scheme, which addresses the segmentation problem in a two-step fashion.
In the first level, we leverage event feature tracking and solve the feature clustering problem via a progressive multi-model fitting pipeline.
In the second level, the event clustering problem is formulated as inferring a 3-D MRF and solved using a spatio-temporal graph-cut method.
The sequentially cascaded scheme leads to a more accurate and efficient event-based motion segmentation, which cannot be achieved by any of the components alone.
The extensive evaluation demonstrates the versatility of the proposed method.

\bibliographystyle{IEEEtran} %

\end{document}